\def\year{2021}\relax
\documentclass[letterpaper]{article} 
\usepackage{aaai21}  
\usepackage{times}  
\usepackage{helvet} 
\usepackage{courier}  
\usepackage[hyphens]{url}  
\usepackage{graphicx} 
\usepackage{natbib}  
\usepackage{caption} 
\usepackage[utf8]{inputenc} 
\usepackage{url}            
\usepackage{booktabs}       
\usepackage{amsfonts}       
\usepackage{amsmath}
\usepackage{amssymb}
\usepackage{graphicx}
\usepackage{subcaption}
\usepackage{nicefrac}       
\usepackage{microtype}      
\usepackage{tikz}
\usepackage{todonotes}

\newcommand{\mtx}[0]{\mathbf}  
\newcommand{\vc}[0]{\boldsymbol}  

\newcommand{\reals}[0]{\mathbb R}

\newcommand{\expectation}[2]{\langle {#1} \rangle_{#2}}
\newcommand{\kldivergence}[2]{\textrm{KL} \left( {#1} || {#2} \right)}

\newcommand{\scriptD}[0]{\mathcal D}
\newcommand{\scriptF}[0]{\mathcal F}
\newcommand{\scriptG}[0]{\mathcal G}

\newcommand{\scriptP}[0]{\mathcal P}

\title{Discovery of Physics and Characterization of Microstructure from Data with Bayesian Hidden Physics Models}
\author{
    Steven Atkinson,\textsuperscript{\rm 1} 
    Yiming Zhang,\textsuperscript{\rm 1} 
    Liping Wang\textsuperscript{\rm 1} \\
}
\affiliations{
    \textsuperscript{\rm 1}GE Research\\
    1 Research Circle\\
    Niskayuna, New York 12309\\
    \{steven.atkinson1, yiming.zhang, wangli\}@ge.com
}
\begin{document}

\maketitle

\begin{abstract}
	There has been a surge in the interest of using machine learning techniques to assist in the scientific process of formulating knowledge to explain observational data.
	We demonstrate the use of Bayesian Hidden Physics Models to first uncover the physics governing the propagation of acoustic impulses in metallic specimens using data obtained from a pristine sample.
	We then use the learned physics to characterize the microstructure of a separate specimen with a surface-breaking crack flaw.
	Remarkably, we find that the physics learned from the first specimen allows us to understand the backscattering observed in the latter sample, a qualitative feature that is wholly absent from the specimen from which the physics were inferred. The backscattering is explained through inhomogeneities of a latent spatial field that can be recognized as the speed of sound in the media.
\end{abstract}

\section{Introduction}
\noindent There is a workflow found ubiquitously throughout many disciplines at the intersection of science and engineering.
It begins when one observes some novel phenomenon or behavior.
One then designs a laboratory experiment to collect measurements to observe the phenomenon under a controlled setting.
Given these observations, one infers relevant physical laws from those observations.
Given this new knowledge, one may proceed with constructing physically-accurate computational models of the discovered physics.
With this tool developed, hypothetical simulations may be carried out \textit{in silico} for the purposes of engineering design and analysis at considerably reduced expense.

Perhaps the most challenging element of the pipeline is the discovery of the relevant physics from the laboratory data.
In fact, there has been a surge of popularity in recent years applying black-box data-driven modeling techniques which involve fitting some general function model to the observation data without the structure of a physical model \cite{bilionis2013multi, zhu2018bayesian, atkinson2019structured}.
While such models tend to be highly general, they struggle to give accurate predictions or possess the sort of generalization that enables significant scientific insight.
The former challenge may be somewhat addressed if it is inexpensive to obtain massive amounts of observational data, though convergence may be unsatisfactorily slow due to the large hypothesis space encoded in typical black-box models.
The latter is more challenging, as extrapolation in data-driven models is notoriously suspect.
For cases in which high accuracy and extrapolation are required, a structured approach that incorporates learned physics seems to be the most effective approach.

In this paper, we consider the above questions in the context of characterizing metal specimens given measurements of their mechanical response to acoustic impulses with a primary focus on discovery of physics from laboratory data.
We consider several of approaches, including end-to-end differentiable physical simulations \cite{degrave2019differentiable} and physics-informed neural networks (PINNs) \cite{raissi2019physics}, eventually culminating in an application of the recently-developed Bayesian Hidden Physics Model \cite{atkinson2020bayesian} to learn the physics of the system and apply it in an extrapolatory regime to understand behavior qualitatively different from that exhibited by the data from which the physics were discovered, i.e.\, backscattering due to a surface-breaking crack.
A key goal is to understand this phenomenon by building physical knowledge from limited data and a broad hypothesis space.
Our machine learning approach contrasts with traditional approaches which seek to calibrate a low-dimensional parameter set for known physics, which typically require considerable domain expertise.
Rather, we would like to be led to the appropriate explanation by the data, as one might need to do in the absence of expert knowledge or in a completely novel setting where such knowledge does not yet exist.

The contributions of this paper are as follows:
\begin{itemize}
	\item We look at the inverse problem on pristine and cracked specimens through both PINNs and a traditional end-to-end differentiable solver that enforces exact agreement on the physics by design. 
	We see that the two qualitatively disagree on their characterization of the cracked specimen, prompting the hypothesis that the wave equation is deficient and that the soft agreement between the wave speed field and the deflection field in PINNs is hiding some missing physics.
	\item We use the technique of Bayesian Hidden Physics Models first demonstrated in \cite{atkinson2020bayesian} to discover a novel differential equation that better describes the data from a hypothesis space containing the wave equation as a special case.
	\item We demonstrate that the physics learned from the pristine specimen may be transferred to explain the presence of backscattering in a specimen with a flaw, demonstrating the data efficiency and validity in extrapolation conferred by a physics-informed approach.
\end{itemize}
The significance of this work is the demonstration that Bayesian Hidden Physics Models may be fruitfully applied to discover physics from real-world data sets, suggesting that the end-to-end scientific workflow described above may be realized.

\subsection{Problem statement}
Consider a physical system with a scalar spatiotemporal observable in two-dimensional space represented as a function $u(x, y, t)$.
First, we assume that the evolution of $u$ is described by a dynamical partial differential equation of the form
\begin{equation}
	u_{tt} = \scriptF[u; \vc \theta_\scriptF]~\textrm{in}~\Omega_{st}=\Omega_s \times [0,T),
	\label{eqn:problem-statement:pde}
\end{equation}
where $\vc \theta_\scriptF$ denotes the (potentially infinite-dimensional) parameterization of the unknown operator $\scriptF$.
The initial-boundary conditions accompanying Eq.\ (\ref{eqn:problem-statement:pde}) are generally not known.
In addition to Eq.\ (\ref{eqn:problem-statement:pde}), one also has access to some finite set of observation data on $u$, denoted as $\scriptD = \{x^{(i)}, y^{(i)}, t^{(i)}, \hat u^{(i)} \}_{i=1}^n$.
Third, we assume that one has access to a simulation code capable of solving Eq.\ (\ref{eqn:problem-statement:pde}).
The simulation code may require the user to specify initial-boundary conditions with parameters $\{\vc \theta_{IC}, \vc \theta_{BC}\}$.
The simulation code may also approximate $\vc \theta_\scriptF$ with some internal, finite-dimensional representation $\vc \theta_f$.
In sum, the parameters of the simulation code are denoted $\vc \theta = (\vc \theta_f, \vc \theta_{IC}, \vc \theta_{BC})$.
Assume that the output of the simulator may be queried as 
$\vc u = \left(u(x^{(1)}, y^{(1)}, t^{(1)}; \vc \theta), \dots, u(x^{(n)}, y^{(n)}, t^{(n)}) \right) \in \reals^n$.
The goal of the simulator calibration problem with known physics is to determine optimal values for theta $\vc \theta$ such that optimally explain $\scriptD$ in some sense.

In the traditional setting, optimality may be defined in terms of some loss function to be minimized such as mean squared error.
However, even in the presence of a large volume of observation data, its coverage in the space relevant to the input of the differential operator may be sparse \cite{atkinson2020bayesian}, yielding many potential parameter sets $(\vc \theta_f)$ consistent with $\scriptD$.
Furthermore, measurements in $\scriptD$ may be noisy.
To address these challenges, we will consider a Bayesian approach by defining a prior $p(\vc \theta_f)$, a likelihood associated with the observed data $p(\hat u|u; \vc \theta_{\hat u})$ with parameters $\vc \theta_{\hat u}$, and use variational inference \cite{blei2017variational} to approximate its posterior by maximizing an evidence lower bound, 
\begin{equation}
	\begin{aligned}
		ELBO(\vc \theta; \scriptD)
		= 
		&\expectation{p(\hat u|u(\vc \theta); \vc \theta_{\hat u})}{q(\vc \theta_f|\vc \theta_{\setminus f})} 
		\\
		&- \kldivergence{q(\vc \theta_f| \vc \theta_{\setminus f})}{p(\vc \theta_f | \vc \theta_{\setminus f})},
	\end{aligned}
	\label{eqn:problem-statement:elbo}
\end{equation}
where $\vc \theta_{\setminus f} = (\vc \theta_{IC}, \vc \theta_{BC}, \vc \theta_{\hat u})$ includes all parameters over which a point estimate will be taken.
Note that while this approach might be extended to $\vc \theta_{IC}$ $\vc \theta_{BC}$, and $\vc \theta_{\hat u}$, in practice the uncertainty associated with these are negligible compared to that on $\vc \theta_f$ if many data are available \cite{atkinson2020bayesian}.

\section{Methodology}
\label{sec:methodology}
Here, we review the methodologies used for microstructure characterization.
We first discuss the acquisition of the laboratory dataset subject to analysis.
We then discuss how characterization may be solved as an inverse problem via gradient-based calibration through a differentiable simulation code or the method of physics-informed neural networks for inverse problems.
Finally, we review Bayesian Hidden Physics Models for discovery of nonlinear partial differential operators with uncertainty quantification as a means of solving the inverse problem without knowing \textit{a priori} what the governing physics are.

\subsection{Data acquisition}
We base our analysis on datasets from two specimens.
The first, referred to as the ``pristine'' specimen, is a nickel polycrystalline alloy.
The second, also used in \cite{shukla2020physics} and referred to as the ``cracked'' specimen, is a National Institute of Standards and Technology (NIST) surface-breaking crack reference standard (RM 8458) in a $7~\textrm{cm} \times 7~\textrm{cm} \times 2~\textrm{cm}$ block of 7075-T651 Aluminum allow substrate material.
To generate the datasets, a piezo-electric transducer imparts a wave packet with a frequency of 5 MHz into the specimen, causing a wavefield to travel across the surface of the specimen.
For the cracked specimen, the wave packet is propagated normal to the crack line.
The wavefield is measured via laser doppler vibrometry at a temporal resolution of 50 MHz (0.02 $\mu$s per sample) and spatial resolution of $0.05~\textrm{mm}$.
For the pristine specimen, the dataset comprises $1024$ measurements in time, on a grid of $400 \times 260$ spatial locations.
For the cracked specimen, the dataset comprises $1024$ measurements in time, on a grid of $240 \times 240$ spatial locations.

\subsection{Modeling approaches}
We now discuss the various modeling approaches we apply to understand the data described above.
A key desideratum is that the relationship between the representation of the microstructure in our models and the resultant measurement field (sampled at a finite set of discrete points) be differentiable.
Therefore, we opt for a heterogeneous continuum model of the specimen.
it is known that, under the assumption that the physics are represented by the (two-dimensional) wave equation,
\begin{equation}
	u_{tt}(x,y,t) = v^2(x, y) \Delta u(x,y,t),
	\label{eqn:wave}
\end{equation}
the presence of microstructure flaws may be modeled as a locally-depressed value of $v$ \cite{seidl2018full}.
Our construction reflects the physical insight that flaws in metallic media may be captured as local domains in which the constitutive properties deviate significantly from their values in the undamaged material.

Another way of modeling a flaw such as a crack is through the imposition of appropriate boundary conditions at the edge of the flaw preventing the transmission of force across it.
However, the use of boundary conditions will be problematic from a practical standpoint because it precludes the use of gradient-based techniques for calibration.
Given the high dimensionality of the simulation parameter space incurred by using implicit neural representations of the parameters of interest, gradient-based techniques are required to make the problem computationally tractable as there are no known gradient-free methodologies with competitive performance for high-dimensional optimization.

\subsubsection{Finite difference physics solver}
We consider the use of a simulation-based calibration scheme built on a finite-difference (FD) implementation of a partial differential equation of the form
\begin{equation}
	u_{tt} = f(u, u_x, u_y, u_{xx}, u_{yy}; \vc \theta_f).
\end{equation}
For the wave equation, 
\begin{equation}
	f = v^2(x, y; \vc \theta_v) (u_{xx} + u_{yy}),
	\label{eqn:waveop}
\end{equation}
The required spatial derivatives are computed via a second-order central finite difference scheme.
We use PyTorch \cite{paszke2019pytorch} to implement $f$ as a convolution operator with fixed kernel matrix.
Implementing the solver completely in PyTorch also results in the ability to easily incorporate neural representations of $v$ and (Dirichlet) boundary conditions \cite{hoyer2019neural, sitzmann2020implicit, raissi2019physics} as well as to differentiate through the simulation via backward-mode automatic differentiation (backpropagation).
While our implementation uses na\"ive automatic differentiation through the whole simulation similarly to \cite{degrave2019differentiable}, more memory-efficient differentiable solvers have been devised utilizing adjoint equations \cite{de2018end} which could be used to relieve the memory cost associated with scaling this method to finer-resolution simulations, though the time cost would not be significantly altered.
Empirically, we find that the latter constitutes the tighter constraint for our problem.

\subsubsection{Physics-informed neural networks for solving inverse problems}
The method of physics-informed neural networks is explained in detail in \cite{raissi2019physics}.
For physics-informed neural networks for inverse problems (henceforth referred to as iPINNs), one parameterizes the solution $u(\cdot)$ to a differential equation with a neural network and supervises the network with the squared residual of the relevant partial differential equation as well as a set of observations on $\hat u$.
This method uses gradient-based optimization to simultaneously optimize the parameters of the solution network as well as any other unknown parameters of the differential equation.

In the case of the current problem, \cite{shukla2020physics} define a solution net $u(x, y, t)$ and a latent field net $v(x,y)$ and supervise both on the heterogeneous wave equation of Eq.\ (\ref{eqn:wave}) as well as $\scriptD$ via a squared error loss on both.
For completeness, we implement their method as a baseline for inversion when the physics are assumed to be known (i.e.\ Eq.\ (\ref{eqn:waveop}) is used with only $v(x,y)$ subject to learning).

\subsubsection{Bayesian Hidden Physics Models}
\label{sec:method:bhpm}
Bayesian Hidden Physics Models (BHPMs) \cite{atkinson2020bayesian} are compositional probabilistic models representing data from multiple experiments and their common underlying physics.
The key differences between BHPMs and inverse PINNs are that the physics are not assumed to be known \textit{a priori} (up to some calibration parameters).
The BHPM approach also extends the Deep Hidden Physics Models approach \cite{raissi2018deep} by quantifying the epistemic uncertainty in the discovered partial differential operator, providing a quantitative answer to the informal question ``What do the data tell us about the physics, and what \textit{don't} they tell us?''

Informally, BHPMs comprise ``leaf'' modules which are in charge of modeling information specific to a single experiment, while the root combines the functional representations of the leaves to learn a common physical governing law.
Formally, let $u: \reals^{d_s+1} \rightarrow \reals$ represent a scalar observable field in $d_s$-dimensional space and time as an implicit parameterization of the measurements in $\scriptD$.
Furthermore, $u$ is assumed to obey some implicit relationship $\Phi[u(\cdot)]=0$.
In the context of the current problem, each leaf contains two functions: $u(x, y, t)$ and $a(x, y)$, both of which are parameterized by multilayer perceptrons (MLPs) with sine activation functions.
We also constrain $a(\cdot) > 0$ by using an element-wise exponential activation function as its final layer.
The former captures the measured deflection in space and time, and the latter is a latent field for which no observations are available.
The two are related through the following differential equation:
\begin{equation}
	u_{tt} = a^2 (x, y) f(u, u_x, u_y, u_{xx}, u_{yy}),
	\label{eqn:BHPM:physics-nonsymmetric}
\end{equation}
Equation (\ref{eqn:BHPM:physics-nonsymmetric}) may be thought of as a generalized wave equation; one recovers the familiar linear wave equation when $a = v(x,y)$ and $f = u_{xx} + u_{yy}$, wherein $a$ gains the physical significance of being the speed of sound in the medium.
Imposing rotational invariance on $f(\cdot)$ we obtain
\begin{equation}
	u_{tt} = a^2 (x, y) f(u, |\nabla u|, \Delta u),
	\label{eqn:BHPM:physics}
\end{equation}
We define the input to $f$ as $\vc \psi = (u, |\nabla u|, \Delta u)$ for brevity.

Finally, reflecting our ignorance about the functional form of the physics, we place a Gaussian process prior over $f(\cdot)$:
\begin{equation}
	f(\cdot) \sim \scriptG \scriptP \left( \mu(\cdot), k(\cdot, \cdot) \right),
\end{equation}
where $\mu(\cdot)$ is a linear function and $k$ is an exponentiated quadratic parametric kernel with automatic relevance determination \cite{rasmussen2006gaussian}.

Decomposing the right hand side of Eq.\ (\ref{eqn:BHPM:physics}) is advantageous because it allows us to separate 
the properties of the particular microstructure of the specimen being studied 
from 
the general physical law; 
the former is represented by $a$, while the latter is captured in $f$ and may be reused once learned.
Furthermore, the multiplicative interaction between $a^2$ and $f$ is advantageous because it encourages an interpretable structure between the various components of the physics while the expressiveness of $a$ and $f$ allow us to discover rich interactions.

The structure of the right hand side of Eq.\ (\ref{eqn:BHPM:physics}) implies that the product $a^2f$ is also \textit{a priori} a Gaussian process-distributed:
\begin{equation}
	a^2 f \sim \scriptG \scriptP \left(a^2 \mu(\cdot), a^4 k(\cdot, \cdot) \right).
\end{equation}
Due to the high volume of data passing through the $f$ as well as the low complexity of the function it is expected to encode\footnote{For example the wave equation is a linear equation in two scalar variables, and even differential equations whose solutions exhibit high complexity come from operators with comparatively simple structure.}, we approximate it using an uncollapsed sparse variational formulation \cite{hensman2013gaussian}, approximating the true posterior $p(f(\cdot) | \scriptD) \approx \int p(f(\cdot) | \vc f_u) q(\vc f_u) d \vc f_u$, where the inducing variables $\vc f_u \in \reals^m$ associated with inducing input locations $\mtx \Psi_u \in \reals^{m \times 3}$ are \textit{a priori} distributed according to $f(\cdot)$ and are given a Gaussian variational posterior.

As in \cite{atkinson2020bayesian}, we equip the measurement and physics targets with Gaussian likelihoods.
Equation (\ref{eqn:problem-statement:elbo}) is maximized with the variational parameters of $f(\cdot)$'s inducing points variables as $\vc \theta_f$ and the likelihood and MLP parameters as $\vc \theta_{\setminus f}$.

\section{Related work}
\label{sec:related-work}
Bayesian calibration of physics codes to data has a long history, with the seminal work of Kennedy and O'Hagan \cite{kennedy2001bayesian}.
Other work \cite{bilionis2013solution} showed how fully Bayesian inversion incorporating probabilistic surrogate models may be facilitated.
More recently, the use of end-to-end differentiable physics engines for inverse problems in control has been demonstrated \cite{de2018end, degrave2019differentiable, rackauckas2020universal}.
Such methods enable the use of gradient-based inference and enable one to calibrate much higher-dimensional parameters, including implicit neural representations of unknown field quantities \cite{sitzmann2020implicit}.
The Bayesian Hidden Physics Models approach \cite{atkinson2020bayesian} combines the Bayesian aspects of earlier work with the powerful neural representations of the latter and can be used to conduct nonparametric inference on nonlinear partial differential operators.

Regarding the application to microstructure characterization, Seidl \cite{seidl2018full} showed that the presence of cracks can be modeled as regions with locally-depressed wave speed.
More recently, Shukla et al.\ \cite{shukla2020physics} showed that the problem of waveform inversion can be accomplished using the method of physics-informed neural networks (PINNs) \cite{raissi2019physics}.
The key difference between our work and these is that we do not specify the physics of the data \textit{a priori} when using Bayesian Hidden Physics Models;
instead, we allow ourselves to discover the physics from the data.
Additionally, we capture the epistemic uncertainty in the physics due to the finite amount of data on which our inference is based, enabling us to identify what knowledge of the physics is still missing and enabling one to be able to carry out a principled adaptive experimental design algorithm.

The crack inversion problem has been tackled traditionally through physics-based finite element simulations. 
Fichtner has elaborated \cite{fichtner2010full} the aspects of full waveform tomography including methods for the numerical solution of the elastic wave equation, the adjoint method, the design of objective functionals and optimization schemes.
Rao et al.\ \cite{rao2016guided} proposed a guided wave tomography method based on full waveform inversion for reconstruction of the remaining wall thickness in isotropic plates. 
The forward model is computed in the frequency domain by solving a full-wave equation in a two-dimensional acoustic model. 
Seidl et al.\ \cite{seidl2018full} elaborated the work to detect structural flaws and their position, dimension and orientation through acoustic full waveform inversion.

\section{Results}
\label{sec:results}
Code implementing the experiments can be found on GitHub.\footnote{
	\url{https://github.com/sdatkinson/BHPM-Ultrasound}
}

\subsection{Calibration of known physics to a pristine specimen}
\begin{figure}[hbt]
	\centering
	\begin{subfigure}[b]{0.45\textwidth}
		\centering
		\includegraphics[width=\textwidth]{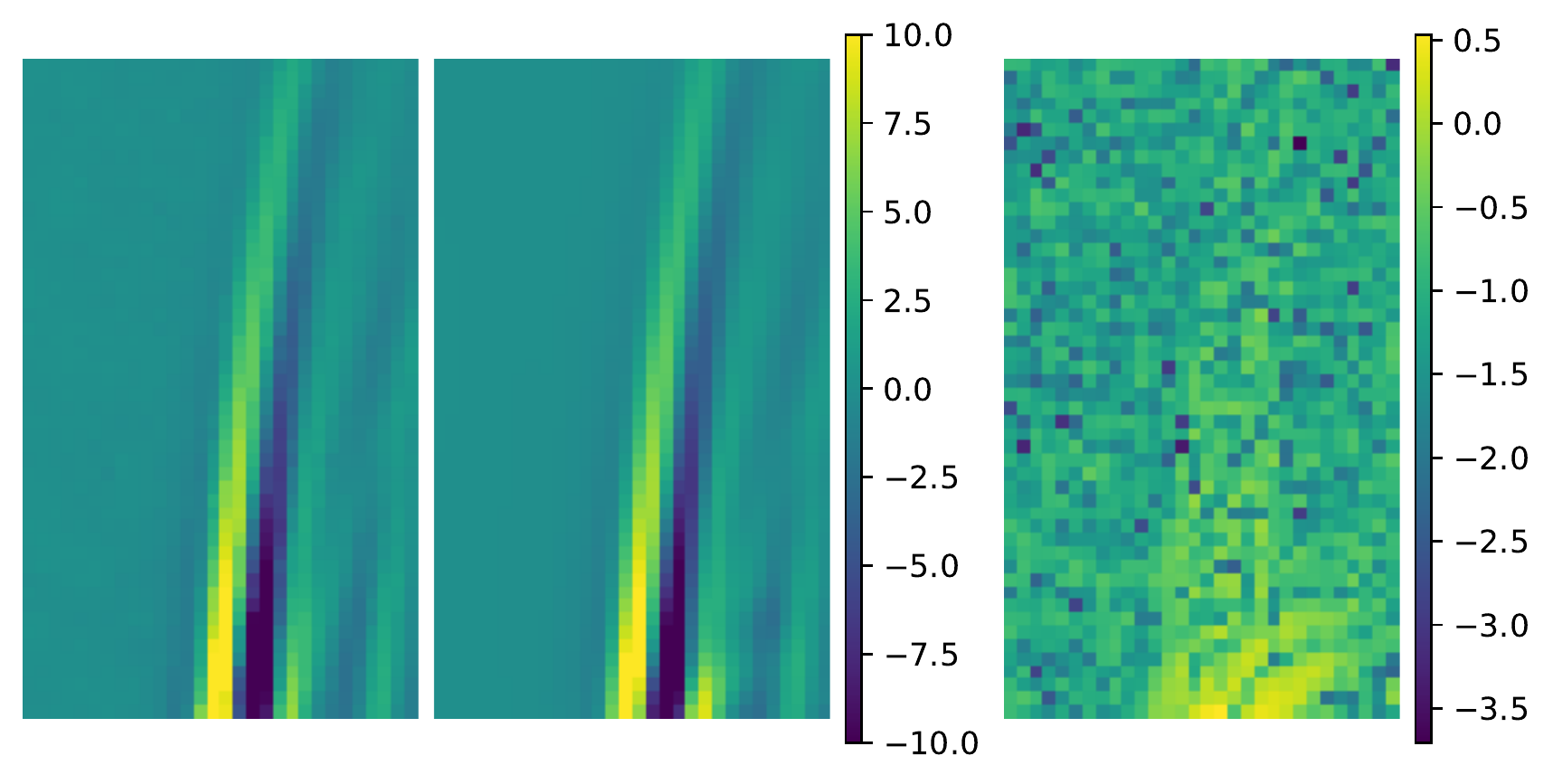}
		\caption{Left to right: $\hat u(\cdot)$, $u(\cdot)$, $\log_{10}|\hat u - u|$.}
		\label{fig:FD:pristine:observations}
	\end{subfigure}
	\\
	\begin{subfigure}[b]{0.15\textwidth}
		\centering
		\includegraphics[width=\textwidth]{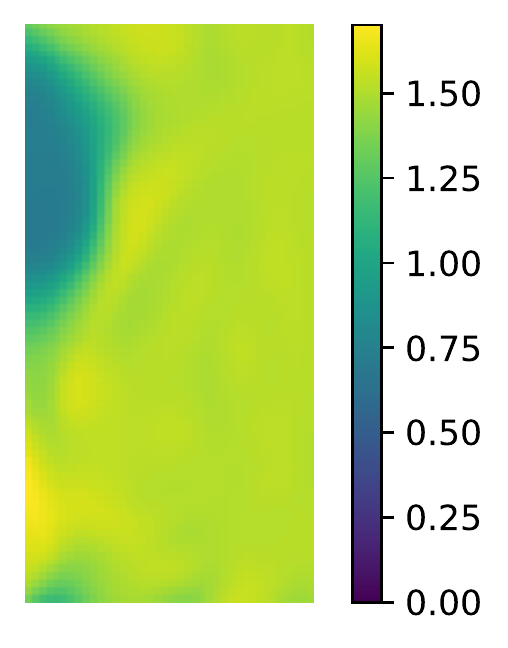}
		\caption{FD, $v(\cdot)$.}
		\label{fig:FD:pristine:a}
	\end{subfigure}
	\caption{
		Calibration of the FD approach to the dataset from the pristine specimen.
		Note that the left side of (\subref{fig:FD:pristine:a}) is unreliable as the wave packet does not travel over it during the window used to calibrate the simulator.
	}
	\label{fig:FD:pristine}
\end{figure}

\begin{figure}[hbt]
	\centering
	\begin{subfigure}[b]{0.45\textwidth}
		\centering
		\includegraphics[width=\textwidth]{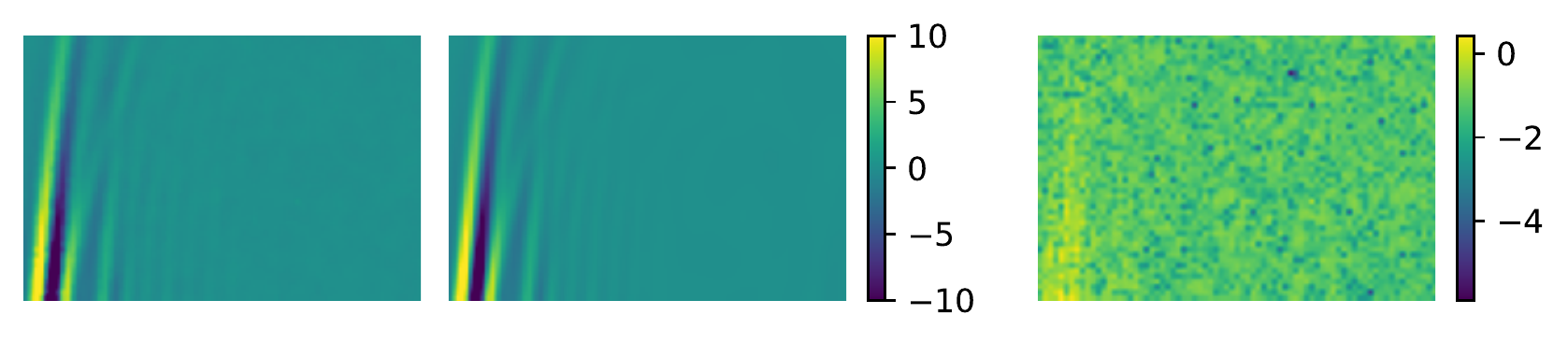}
		\caption{Left to right: $\hat u(\cdot)$, $u(\cdot)$, $\log_{10}|\hat u - u|$.}
		\label{fig:iPINN:pristine:observations}
	\end{subfigure}
	\\
	\begin{subfigure}[b]{0.45\textwidth}
		\centering
		\includegraphics[width=\textwidth]{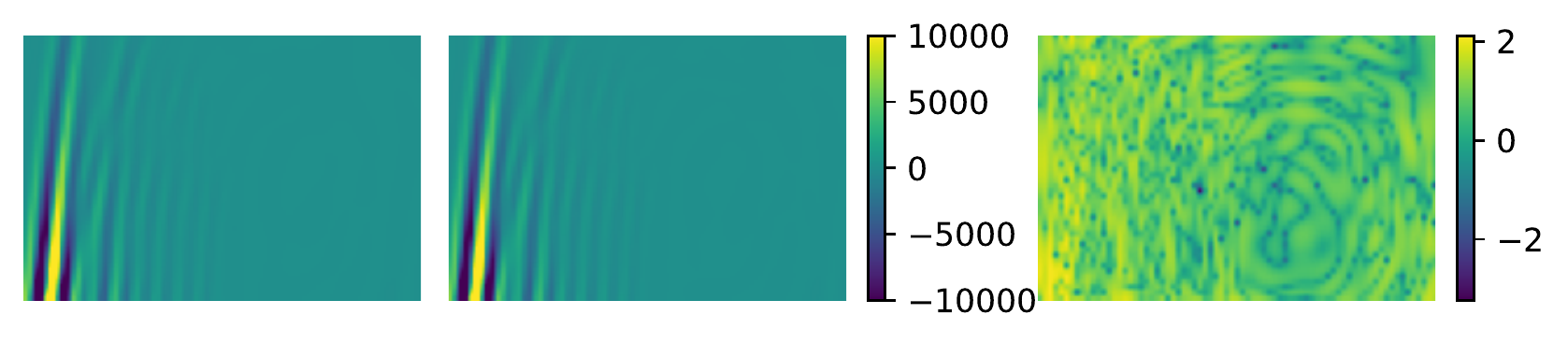}
		\caption{Left to right: $u_{tt}(\cdot)$, $v^2f$, $\log_{10}|u_{tt} - v^2f|$.}
		\label{fig:iPINN:pristine:physics}
	\end{subfigure}
	\\
	\begin{subfigure}[b]{0.15\textwidth}
		\centering
		\includegraphics[width=\textwidth]{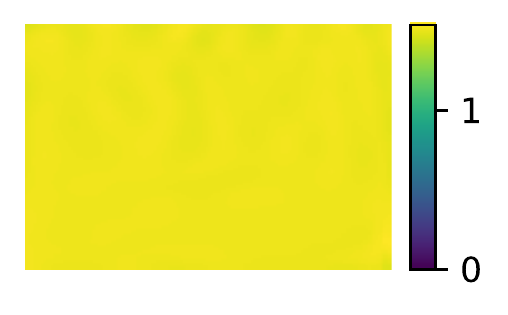}
		\caption{$v(\cdot)$.}
		\label{fig:iPINN:pristine:a}
	\end{subfigure}
	\caption{iPINN learned models for the observations and physics of the pristine specimen.}
	\label{fig:iPINN:pristine}
\end{figure}

We first demonstrate the various modeling approaches on modeling the ``pristine'' specimen.
Figure \ref{fig:FD:pristine} shows a snapshot in time of the results calibrating the FD approach to data cropped to a 
$1.5~\textrm{mm} \times 2.5~\textrm{mm} \times 1~\mu\textrm{s}$ window, while Fig.\ \ref{fig:iPINN:pristine} shows the results for the iPINN approach applied to a 
$3.75~\textrm{mm} \times 2.5~\textrm{mm} \times 3~\mu\textrm{s}$ window.
For the FD approach we report the agreement between simulated and observed displacement fields as well as the inferred $v(\cdot)$.
For the iPINN approach we additionally report on the residual of the physics equation since it is not enforced by construction.
We see that both are able to reproduce the observed data and explain the data in terms of the wave equation physics that they assume.

\subsection{Calibration of known physics to a cracked specimen}
We now proceed to characterize a specimen with a localized flaw.
We expect that this will show up as a corresponding region in which $v$ is noticeably lower.
This problem is challenging due to the details in the geometry of the flaw.

\begin{figure}[hbt]
	\centering
	\begin{subfigure}[b]{0.45\textwidth}
		\centering
		\includegraphics[width=\textwidth]{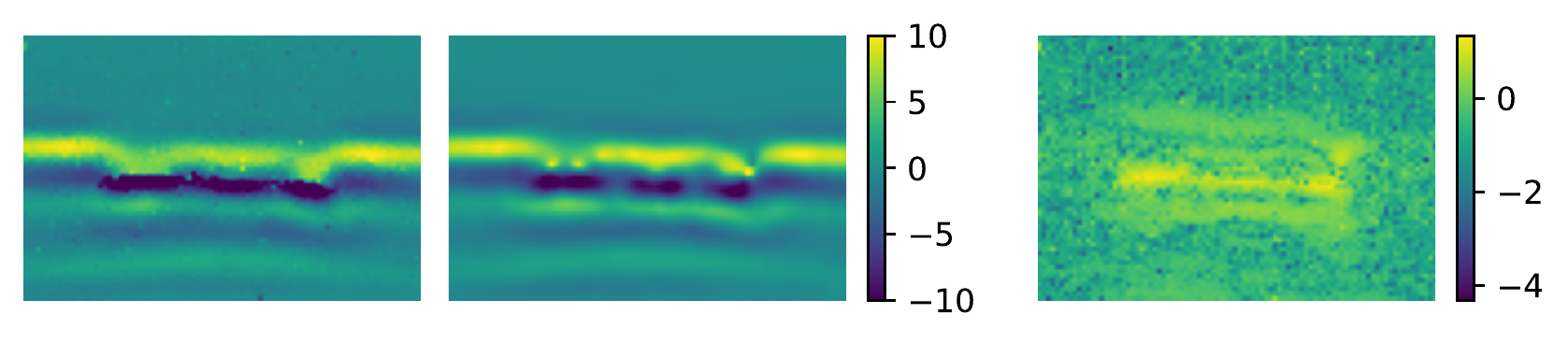}
		\caption{Left to right: $\hat u(\cdot)$, $u(\cdot)$, $\log_{10}|\hat u - u|$.}
		\label{fig:FD:cracked:observations}
	\end{subfigure}
	\\
	\begin{subfigure}[b]{0.15\textwidth}
		\centering
		\includegraphics[width=\textwidth]{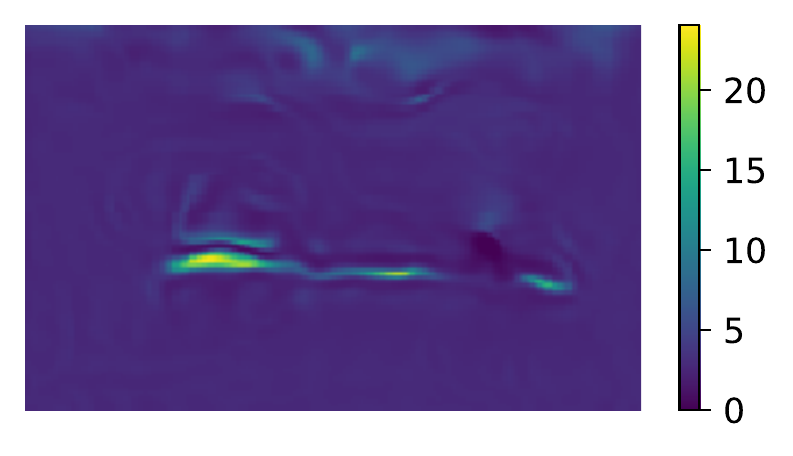}
		\caption{$v(\cdot)$.}
		\label{fig:FD:cracked:a}
	\end{subfigure}
	\caption{Calibration of the FD solver to the cracked specimen.}
	\label{fig:FD:cracked}
\end{figure}

\begin{figure}[hbt]
	\centering
	\begin{subfigure}[b]{0.45\textwidth}
		\centering
		\includegraphics[width=\textwidth]{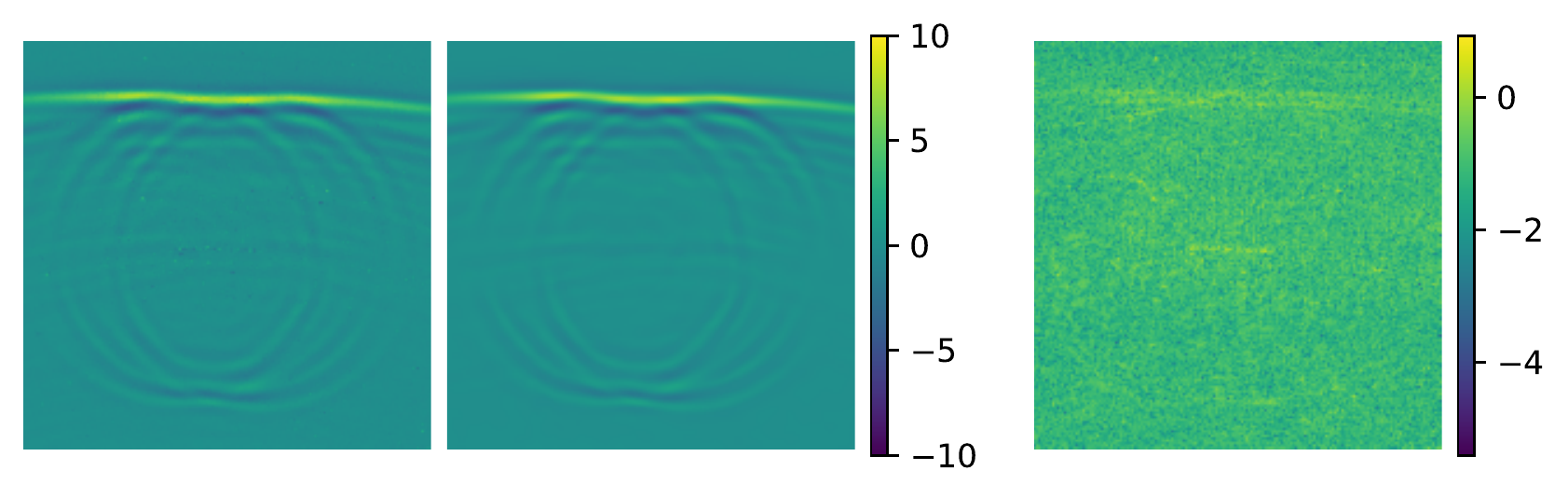}
		\caption{Left to right: $\hat u(\cdot)$, $u(\cdot)$, $\log_{10}|\hat u - u|$.}
		\label{fig:iPINN:cracked:observations}
	\end{subfigure}
	\\
	\begin{subfigure}[b]{0.45\textwidth}
		\centering
		\includegraphics[width=\textwidth]{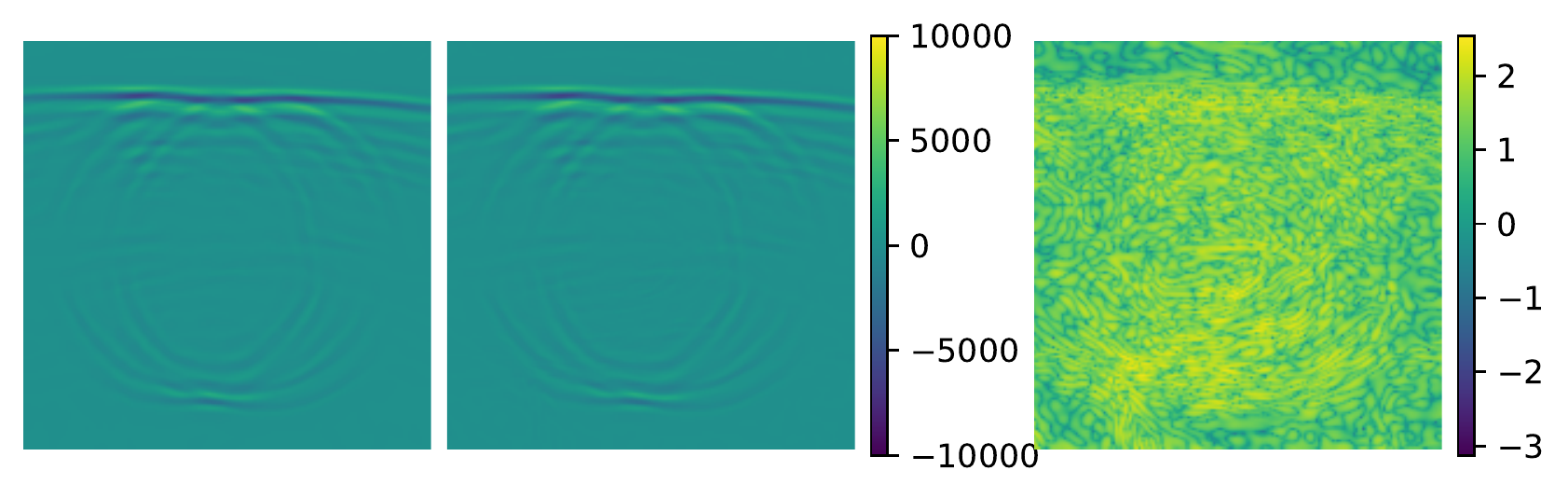}
		\caption{Left to right: $u_{tt}(\cdot)$, $v^2f$, $\log_{10}|u_{tt} - v^2f|$.}
		\label{fig:iPINN:cracked:physics}
	\end{subfigure}
	\\
	\begin{subfigure}[b]{0.15\textwidth}
		\centering
		\includegraphics[width=\textwidth]{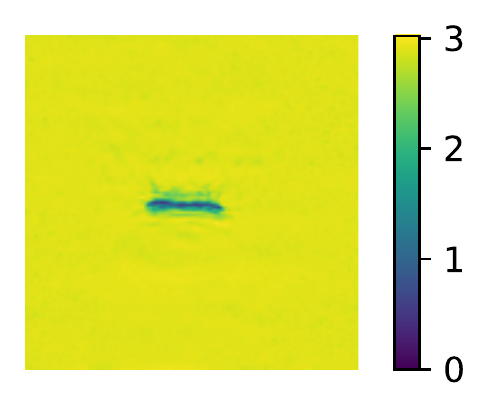}
		\caption{$v(\cdot)$.}
		\label{fig:iPINN:cracked:a}
	\end{subfigure}
	\caption{iPINN learned models for the observations and physics of the cracked specimen.}
	\label{fig:iPINN:cracked}
\end{figure}

Figure \ref{fig:FD:cracked} shows a snapshot in time of the calibration result using the FD method on the cracked specimen windowed to 
$4.5~\textrm{mm} \times 3~\textrm{mm} \times 1.2~\mu\textrm{s}$, and Fig.\ \ref{fig:iPINN:cracked} shows compares the calibration using the iPINN model with data cropped only in time to $5~\mu\textrm{s}$.
We were not able to calibrate as large an area for the FD solver due to its limitations related to discretization and the computational cost associated with the method.\footnote{The FD approach parameters were selected to limit it to taking at most about an order of magnitude more time than the other approaches.}

Comparing the FD and iPINN approaches, we see that there are some discrepancies in the neighborhood of the flaw.
On one hand, the relationship between $u$ and $v$ is enforced to solver precision in the FD approach. 
By contrast, agreement in the physics is only enforced through a soft constraint in the case of the iPINN approach.
Based on the observations that (1) the FD solver struggles to accurately reproduce the data in the vicinity of the crack and (2) that the physics are least in agreement in the vicinity of the crack for the iPINN approach, we infer that there may be missing physics in the standard wave equation that preclude its direct application to the case of flawed specimens.
In the next sections, we explore this hypothesis by learning the physics from data using a Bayesian Hidden Physics Model with a larger physics hypothesis space rather than calibrating the postulated wave equation to data.

\subsection{Discovery of unknown physics from a pristine specimen}
\begin{figure}[hbt]
	\centering
	\begin{subfigure}[b]{0.45\textwidth}
		\centering
		\includegraphics[width=\textwidth]{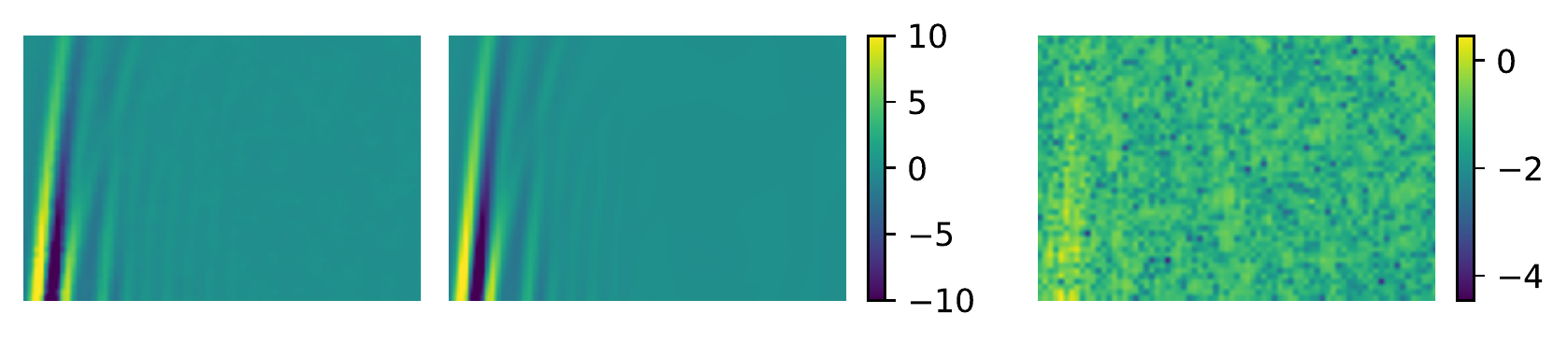}
		\caption{Left to right: $\hat u(\cdot)$, $u(\cdot)$, $\log_{10}|\hat u - u|$.}
		\label{fig:BHPM:pristine:observations}
	\end{subfigure}
	\\
	\begin{subfigure}[b]{0.45\textwidth}
		\centering
		\includegraphics[width=\textwidth]{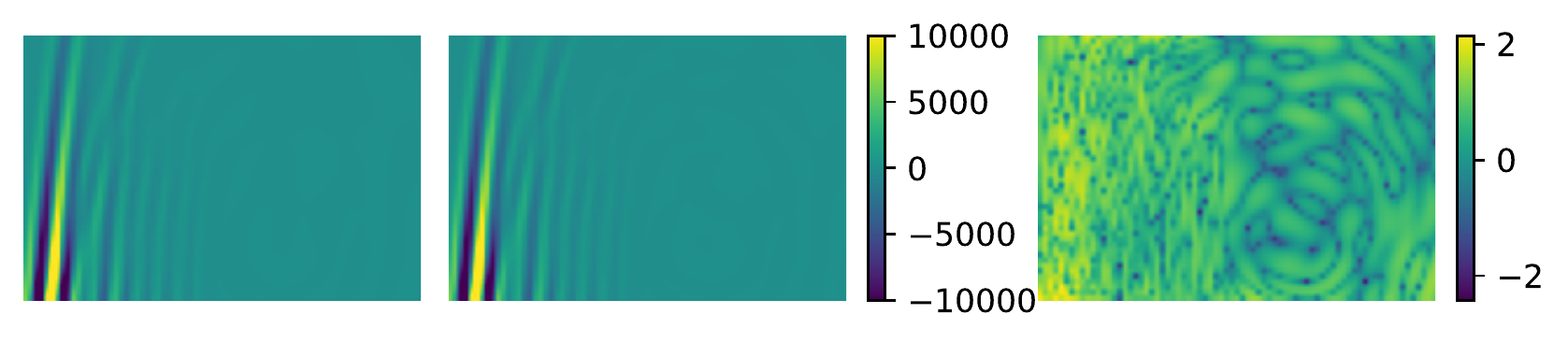}
		\caption{Left to right: $u_{tt}(\cdot)$, $a^2f$, $\log_{10}|u_{tt} - a^2f|$.}
		\label{fig:BHPM:pristine:physics}
	\end{subfigure}
	\\
	\begin{subfigure}[b]{0.15\textwidth}
		\centering
		\includegraphics[width=\textwidth]{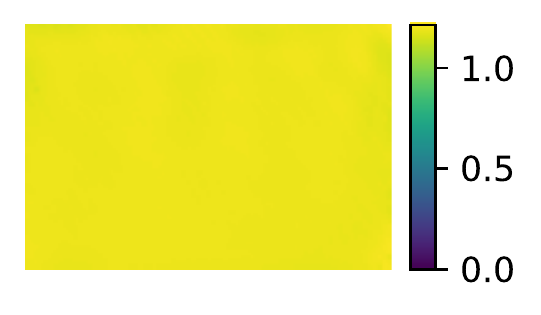}
		\caption{$a(\cdot)$.}
		\label{fig:BHPM:pristine:a}
	\end{subfigure}
	\caption{BHPM learned models for the observations and physics of the pristine specimen.}
	\label{fig:BHPM:pristine}
\end{figure}

A Bayesian Hidden Physics model is constructed as discussed above and trained on data obtained from the pristine sample windowed to 
$3.75~\textrm{mm} \times 2.5~\textrm{mm} \times 3~\mu\textrm{s}$.
Figure \ref{fig:BHPM:pristine} shows the agreement obtained for the observation data and discovered physics.
The root mean squared error for the observation function is 
$1.35 \times 10^{-1}$ 
and the root mean squared residual of the learned physics using the posterior mean of $f$ is 
$1.44 \times 10^1$.

\subsection{Characterization of flawed specimen using learned physics}
\begin{figure}[hbt]
	\centering
	\begin{subfigure}[b]{0.45\textwidth}
		\centering
		\includegraphics[width=\textwidth]{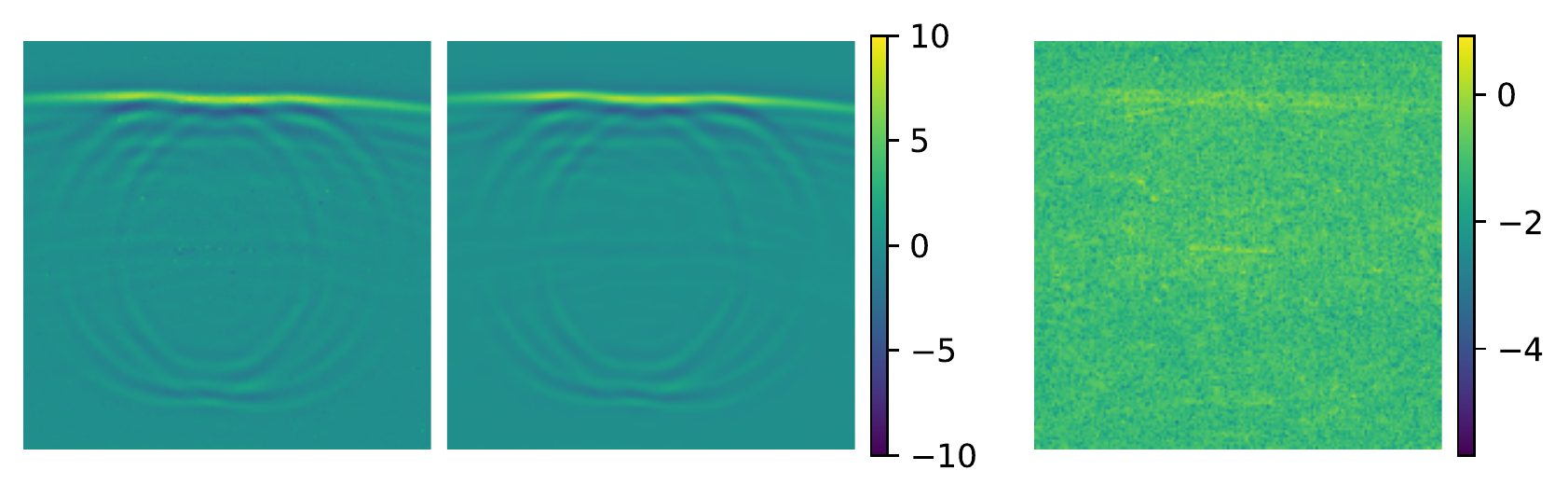}
		\caption{Left to right: $\hat u(\cdot)$, $u(\cdot)$, $\log_{10}|\hat u - u|$.}
		\label{fig:BHPM:cracked:observations}
	\end{subfigure}
	\\
	\begin{subfigure}[b]{0.45\textwidth}
		\centering
		\includegraphics[width=\textwidth]{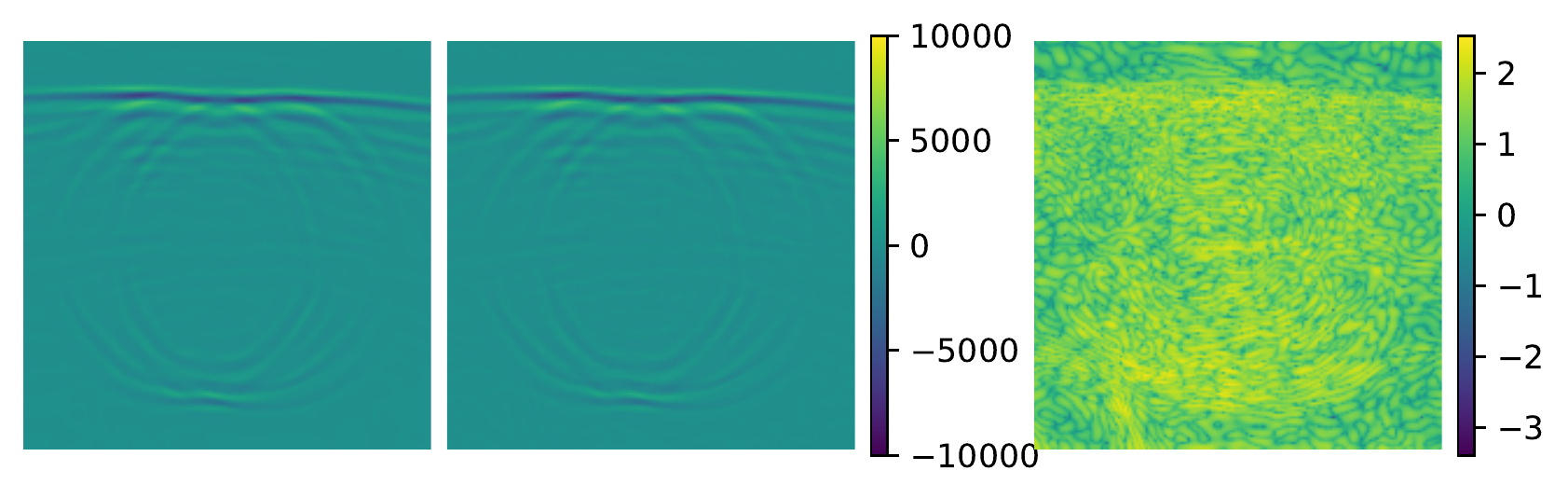}
		\caption{Left to right: $u_{tt}(\cdot)$, $a^2f$, $\log_{10}|u_{tt} - a^2f|$.}
		\label{fig:BHPM:cracked:physics}
	\end{subfigure}
	\\
	\begin{subfigure}[b]{0.15\textwidth}
		\centering
		\includegraphics[width=\textwidth]{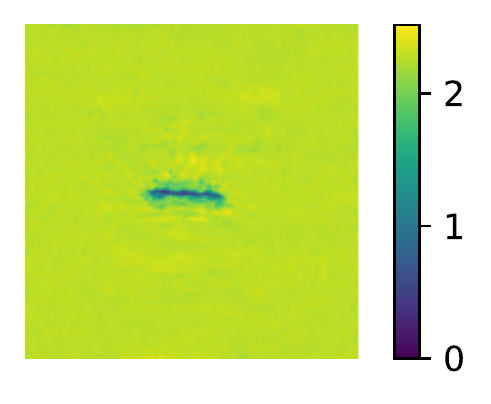}
		\caption{$a(\cdot)$.}
		\label{fig:BHPM:cracked:a}
	\end{subfigure}
	\caption{BHPM learned models for the observations and physics of the cracked specimen.}
	\label{fig:BHPM:cracked}
\end{figure}

We now transfer the learned physics from the pristine specimen to characterize the cracked specimen, windowed in time to $5~\mu\textrm{s}$.
To do this, we initialize a second BHPM model, using the posterior on $f(\cdot)$ learned from the first experiment.
Only the MLPs for $u$ and $a$ are trained to characterize the sample.
Figure \ref{fig:BHPM:cracked} shows the agreement obtained for the observation data and discovered physics.
The root mean squared error for the observation function is 
$2.38 \times 10^{-1}$ 
and the root mean squared residual of the learned physics using the posterior mean of $f$ is 
$1.46 \times 10^2$.
Remarkably, we see that $a(\cdot)$ takes on depressed values in the vicinity of the crack in the specimen, indicating that the backscattering in the observations can be traced back to the inhomogeneity in the material, while the far-field character is relatively homogeneous.
Note also that the factors $a$ and $f$ comprising the right hand side are not individually uniquely identifiable since an arbitrary factor of $\alpha$ could be multiplied and divided into $a^2$ and $f$, respectively; therefore, the absolute magnitude of $a$ in Figs.\ \ref{fig:BHPM:pristine:a} and \ref{fig:BHPM:cracked:a} is arbitrary and should not be directly compared with $v(\cdot)$. 
However, the relative magnitude within each field as well as the relative magnitude across experiments is meaningful due to the fact that $f$ was held fixed.

\section{Conclusions}  
\label{sec:conclusion}
In this work, we considered the problem of characterizing the microstructure of metal specimens with and without flaws from data via physics discovery.
We began by establishing agreement between two physics-based approaches using a end-to-end differentiable finite difference code as well as the inverse physics-informed neural networks approach.
While both methods exhibited good agreement for pristine specimens, we found that they disagreed more in the case of a specimen with a crack flaw.
Following this observation, we used Bayesian Hidden Physics Models to learn the physics from data from the pristine specimen, allowing for a larger hypothesis space than the standard wave equation.
We then transferred the learned physics to the flawed specimen and characterized its microstructure through the lens of the learned physics.
We found that the BHPM approach is able to identify the flaw in the specimen.

It is of general interest to develop an end-to-end scientific and engineering pipeline wherein one may begin by observing some novel behavior, collect measurements within a laboratory setting, deduce the relevant physical laws from those observations, then proceed with constructing a physically-accurate computational model of the discovered physics such that simulation may be leveraged for engineering design and analysis.
All elements of this pipeline have been reduced to practice in the past.
However, the most challenging element of the pipeline is that of discovery of novel physics from data; this is typically exclusively in the realm of human scientists.
Our results using BHPM demonstrate that probabilistic machine learning techniques may expedite this process.

A number of questions of scientific interest persist, as do several opportunities for further engineering development based on our work.
On the more fundamental side, the probabilistic side of physics discovery through machine learning is in its infancy.
However, the recent advance of the BHPM approach stands to unite the large body of Bayesian inquiry to more open-ended exploration and discovery in physics; in particular, questions of optimal experimental design may now be treated rigorously, as the BHPM approach places questions of physics knowledge with an information-theoretic context.
On the side of engineering development and capability, it is of interest to incorporate the learned physics into novel computational models that can be brought to bear to explore the space of material behaviors under various heterogeneities and external actions.
The construction of physics-based codes and surrogates therein will lead the way to efficient sensing methodologies capable of real-time characterization and monitoring of materials and structures in engineering systems.

\section*{Acknowledgments}
The authors thank the United States Air Force Research Laboratory for providing the data analyzed in this work for public release under Distribution A as defined by the United States Department of Defense Instruction 5230.24. 
This material is based upon work supported by the Defense Advanced Research Projects Agency (DARPA) under Agreement No.\ HR00111990032.

We thank the developers of the software from which our research code development has benefitted, including
H5Py \cite{collette2013python},
JAX \cite{bradbury2018jax},
Matplotlib \cite{hunter2007matplotlib},
NumPy \cite{oliphant2006guide,van2011numpy},
Pandas \cite{mckinney2011pandas},
PyTorch \cite{paszke2019pytorch},
Scikit-learn \cite{pedregosa2011scikit},
SciPy \cite{virtanen2020scipy},
and
tqdm \cite{da2019tqdm}.


\end{document}